\documentclass[10pt]{relaxed_system_lab}
\ifPDFTeX\usepackage[utf8]{inputenc}\fi
\usepackage[T1]{fontenc}
\usepackage{charter}
\usepackage{fontawesome5}
\usepackage{amsmath,amssymb}
\usepackage{array,tabularx}
\usepackage{xspace}
\usepackage{algorithm}
\usepackage{algpseudocode}
\usepackage{tikz}
\usetikzlibrary{arrows.meta,positioning,fit,backgrounds}
\usepackage{xurl}
\usepackage{needspace}
\newcommand{\sys}{\textsc{AReaL-RSI}\xspace}
\newcommand{\cost}{\mathbf{C}}
\newcommand{\budget}{\mathbf{B}}
\newcolumntype{L}[1]{>{\raggedright\arraybackslash}p{#1}}
\newcolumntype{Y}{>{\raggedright\arraybackslash}X}
\setlist{nosep,leftmargin=*}
\renewcommand{\arraystretch}{1.15}

\newtcolorbox{rqsummary}[1]{
  colback=labblue!5,
  colframe=labblue!50,
  boxrule=0.4pt,
  arc=1mm,
  boxsep=0pt,
  left=6pt,right=6pt,top=6pt,bottom=6pt,
  before skip=7pt,after skip=7pt,
  before upper={\textbf{#1.}\enspace}
}

\title{\sys: Adaptive Scaling and Coordination of Recursive Self-Improvement}

\author{Wenhao Zhou$^{1,2}$, Jiajun Song$^1$, Haitao Wang$^1$, Zipeng Qiu$^{1,3}$, Yingqi Peng$^{1,2}$, Ruida Xu$^1$, Jiale Li$^1$, Wenshuang Li$^{1,3}$, Shusheng Xu$^1$, Wei Dong$^2$, Hao Dai$^1$, Tongkai Yang$^1$, Yi Gao$^2$, Binhang Yuan$^{1,3}$}

\affiliation{$^1$Ant Group, $^2$Zhejiang University, $^3$HKUST}

\abstract{
Scaling recursive self-improvement requires additional computation to produce useful research progress in reaching persistent objectives. Intuitively, more researcher agents can explore more hypotheses, but early evaluation scores might misrepresent their promise, shared findings could lose their applicability conditions, and useful procedures could remain confined to completed investigations. In this paper, we present \sys, a framework for organizing a persistent research team of agents without updating backbone LLM weights. We formulate the scaling problem in terms of evaluated outcomes and whole-system resource expenditure. Concretely, a leader agent selects task-appropriate research orientations and guides ongoing investigations; conditional memory sharing distinguishes unresolved problems from supported findings and enables recipient adaptation. During recursive evolution, after each campaign, evidence-based review revises researcher orientation, knowledge, and skills for later campaigns. Experiments compare \sys with \textsc{Codex-Goal} on downstream harness optimization, GPU kernel optimization, and AI research tasks using two backbones. Under the reported resource allocations, \sys improves Terminal Bench performance by 2.2 percentage points for each backbone and increases mean KernelBench speedup by 32.2--36.8\%; further scalability comparisons favor the design of \sys over native scaling: combining experience sharing with specialized researcher roles yields 1.9–27.5\% lower kernel latency than independent exploration and 1.3–33.9\% lower latency than sharing alone across all seven comparable task–backbone pairs. 
}

\begin{document}
\maketitle

\section{Introduction}
\label{sec:intro}

Recursive self-improvement (RSI) seeks to improve an AI agent through repeated self-modification and empirical evaluation~\citep{dgm}. The scope of RSI can extend from improving a task solution to revising the procedures that generate subsequent improvements~\citep{hyperagents,dreamrsi}. This naturally raises a test-time scaling problem, i.e., \textit{how should additional inference and experimental resources improve the outcomes of RSI?} A system can investigate more hypotheses, share discoveries between ongoing investigations, or retain useful procedures for later tasks. In this paper, we study \textit{how a leader agent can coordinate these uses of computation across persistent researcher agents, with an external researcher agent that is open to revision.}

\begin{figure}[t]
    \centering
    \includegraphics[width=\textwidth]{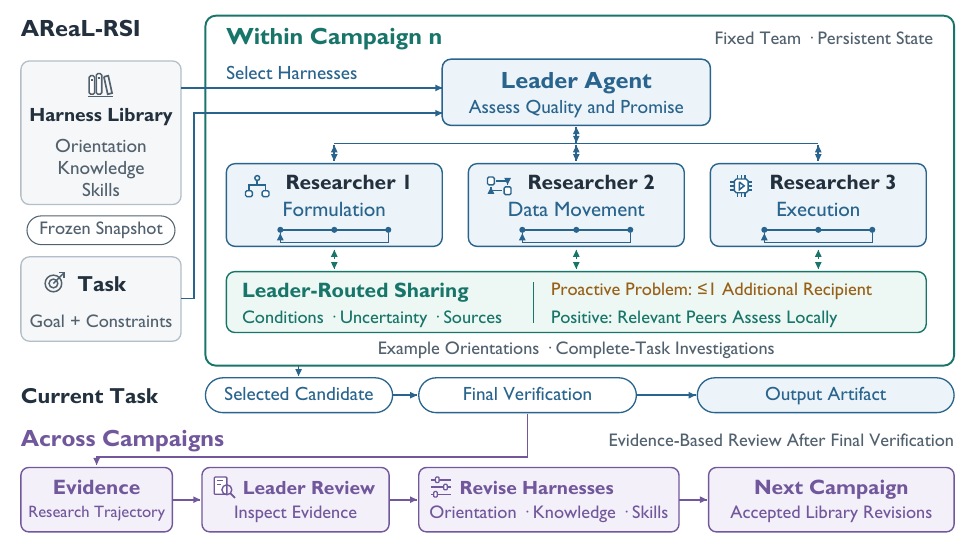}
    \vspace{-1.0em}
    \caption{\sys organizes RSI adaptive scaling. A leader agent selects task-conditioned harnesses and guides persistent researcher agents, each investigating the complete task. Shared experience retains sources, conditions, and uncertainty. After verification, the leader agent reviews the research trajectory and proposes revisions to orientation, knowledge, and skills. Accepted revisions become available to future campaigns.}
    \label{fig:areal-rsi-overview}
    \vspace{-1.0em}
\end{figure}

Scaling out is essential for RSI because, when pursuing persistent objectives, research progress depends on exploring alternatives and testing them, not simply producing a longer answer across recursive iterations. For example, GPU kernel optimization and machine learning engineering require executable artifacts whose quality is established through experimentation~\citep{kernelbench,rebench,mlebench}. Parallel researcher agents can investigate complementary hypotheses while retaining the depth needed to implement and test each one. Sharing a useful diagnosis can then benefit several investigations, and retaining the resulting procedure can reduce repeated work on later tasks. A leader agent realizes this opportunity by combining evidence across the team of research agents to guide investigative priorities and decide which findings deserve wider attention. Comprehensively, the scaling objective is to turn additional research capacity into useful progress while accounting for the cost of coordination and evaluation.

Three key difficulties limit RSI from easily scaling out. \underline{First}, early artifact quality is an incomplete signal of investigative promise: a weak implementation may contain a valuable idea, while an early improvement may depend on a fragile assumption; thus, simply redirecting every researcher agent toward the currently best-scoring direction can, therefore, abandon useful alternatives. \underline{Second}, the experience is conditional: A finding that holds for one implementation or numerical regime may mislead another researcher agent when the conditions of the finding are omitted; broad sharing can also homogenize exploration~\citep{interactiontax}. \underline{Third}, progress on a task does not automatically improve the next research process: a completed investigation may contain reusable diagnostics or experimental procedures, but extracting them requires separating general methods from task-specific artifacts and uncertain explanations. These difficulties connect the leader agent's immediate coordination decisions to the longer-term reuse of research experience.

Existing approaches offer mechanisms for coordination, memory, and procedure evolution, but combining these capabilities still requires a policy for when to share evidence and how to revise reusable research procedures. AI co-scientist and Arbor organize collaborative investigation~\citep{coscientist,arbor}, while CoMem maintains private and collective memory~\citep{comem}. Population-based training and fluid research search adapt the allocation of parallel work~\citep{pbt,efficiencymatters}; AlphaEvolve maintains an evolutionary program archive, and Hyperagents and Dream-RSI revise aspects of the improvement process~\citep{alphaevolve,hyperagents,dreamrsi}. In this paper, our focus is the coupling between evidence interpretation, selective information exchange, and reuse of research procedures to scale out RSI. 

Towards this end, we introduce \sys to organize RSI on two timescales (Figure~\ref{fig:areal-rsi-overview}). Within a campaign (i.e., one execution iteration of \sys on a specified task), a leader agent selects researcher harnesses suited to the task and provides evidence-driven guidance to persistent researcher agents. Conditional memory sharing links findings to their sources and tested conditions, limits proactive propagation of unresolved problems, and leaves recipients responsible for assessing applicability. After a campaign, evidence-based review revises research orientation, reusable knowledge, and operational skills for subsequent campaigns. These mechanisms address the three difficulties through explicit coordination and revision procedures. Our contributions are summarized as follows:

\noindent\textbf{\underline{Contribution 1}}.
We formulate adaptive RSI scaling as a problem of organizing persistent investigations and reusable research procedures under a declared resource allowance (\S\ref{sec:problem}). The formulation distinguishes current artifact quality from investigative promise and separates the opportunity offered by additional computation from the efficiency of research team organization, which connects task-conditioned team composition and leader-agent guidance to adaptation within and across research campaigns.

\noindent\textbf{\underline{Contribution 2}}.
We develop conditional memory sharing (\S\ref{sec:method}) and experience-driven researcher-harness evolution to support productive scaling (\S\ref{sec:execution}). Source-linked findings support selective propagation and recipient adaptation, while post-campaign review converts experimental experience into revisions of research orientation, knowledge, and skills. The mechanisms distinguish reuse among ongoing investigations from changes to the procedures available to future teams.

\noindent\textbf{\underline{Contribution 3}}.
We evaluate \sys against \textsc{Codex-Goal}~\citep{codexgoal} across downstream harness optimization, GPU kernel optimization, and AI research tasks with two LLM backbones (\S\ref{sec:evaluation}). Under the reported resource allocations, \sys increases Terminal Bench resolution by 2.2 percentage points for each backbone and achieves 32.2--36.8\% higher mean KernelBench speedup. Further comparisons of scaling research configurations support \sys’s coordinated scaling design: experience sharing with specialized researcher roles achieves 1.9–27.5\% lower kernel latency than independent exploration and 1.3–33.9\% lower latency than sharing alone across all seven comparable task–backbone pairs.

\section{Adaptive RSI Scaling}
\label{sec:problem}

Scaling RSI requires organizing additional computation into useful investigations. We define the scaling objective, then describe team composition and leader-agent guidance. This organization supports sharing within campaigns and harness revision across campaigns (Figure~\ref{fig:areal-rsi-overview}).

\subsection{Why Scale Out RSI?}
\label{sec:scaling-objective}

Parallel researcher agents can test competing hypotheses under costly or delayed feedback. However, duplicate work and indiscriminate sharing can waste computation or cause premature convergence. Effective scaling requires coordinating investigative priorities and information exchange alongside inference and experiment costs. Retaining useful procedures in the harnesses of a researcher agent creates a second opportunity: effort spent on one task can improve how later tasks are investigated.

\textbf{Problem Formulation.}
A campaign is one execution of \sys on a task $q_n$ with a persistent objective and a fixed incoming harness-library snapshot. Let $n$ index campaigns, $\pi$ denote an admissible organization policy, and $x_n^{\mathrm{out}}(\pi)$ denote the output artifact. The final utility $U_n(x)$ is task-specific, with larger values preferred and invalid outputs assigned the task's failure value. For resource expenditure vector $\cost_n^{\mathrm{total}}(\pi)$ and allowance $\budget_n$, the objective is
\begin{equation}
 \max_{\pi}\;\mathbb{E}\!\left[U_n\!\left(x_n^{\mathrm{out}}(\pi)\right)\right]
 \quad\mathrm{subject\ to}\quad
 \mathbb{E}\!\left[\cost_n^{\mathrm{total}}(\pi)\right]\preceq\budget_n.
 \label{eq:scaling-objective}
\end{equation}
The expectation covers stochastic research and evaluation; $\preceq$ imposes componentwise resource constraints. The policy governs team selection, guidance, and sharing. Costs include researcher-agent execution, leader-agent inference, experiments, and post-campaign review (Equation~\ref{eq:total-cost}); review benefits are assessed in later campaigns. This per-campaign objective defines a comparison rather than a runtime optimizer. 

\subsection{Task-Conditioned Team Composition}
\label{sec:composition}
\label{sec:resources}

Each campaign uses a fixed team of $K$ researcher agents. The task $q_n$ specifies the objective, admissible artifacts, constraints, and evaluation interface. Every researcher agent investigates the complete task and retains local state across interactions. Researcher harnesses provide complementary orientations, knowledge, and skills (\S\ref{sec:harness}).

\textbf{Harness Selection.}
At campaign creation, the system snapshots the harness library as $\overline{\mathcal{L}}_n$. The leader agent selects harness assignments according to the task's constraints, unresolved questions, and verification cost:
\begin{equation}
 (h_{n,1},\ldots,h_{n,K})
 \sim \pi_{\mathrm{select}}(\,\cdot\mid q_n,\overline{\mathcal{L}}_n),
 \qquad h_{n,i}\in\overline{\mathcal{L}}_n.
 \label{eq:composition}
\end{equation}
Here $h_{n,i}$ is researcher agent $i$'s harness and $\pi_{\mathrm{select}}$ is the leader agent's selection policy. Multiple researcher agents may receive the same harness when an orientation merits additional exploration. Assigned versions remain fixed within the campaign, although local working files remain editable. Guidance adjusts investigative focus without replacing team members or dividing responsibility into disjoint subtasks.

\subsection{Evidence-Driven Coordination}
\label{sec:coordination}
\label{sec:assessment}
\label{sec:allocation}

The leader agent uses progress reports and experimental records to guide the team. At coordination event $t$, let $\mathcal{E}_{n,t}$ denote the available public evidence and $g_{n,t}$ the leader agent's context: inspected evidence, the current plan, and reported resource availability. Actions follow
\begin{equation}
 a^{\mathrm{L}}_{n,t}
 \sim \pi_{\mathrm{coord}}(\,\cdot\mid q_n,g_{n,t}),
 \label{eq:coordination}
\end{equation}
where $\pi_{\mathrm{coord}}$ uses model reasoning and tools to request evidence, revise the plan, advise a researcher agent, or await feedback.

\textbf{Quality vs Promise.}
Coordination distinguishes \emph{artifact quality} from \emph{investigative promise}. Let $\mathcal{X}^{\mathrm{ok}}_{n,t}$ contain saved, non-withdrawn candidates that passed the available checks, and let $s_n(x)$ be their recorded score, oriented so that larger values are better. For a nonempty candidate set, the incumbent is
\begin{equation}
 x^{\star}_{n,t}\in
 \operatorname*{arg\,max}_{x\in\mathcal{X}^{\mathrm{ok}}_{n,t}}s_n(x).
 \label{eq:incumbent}
\end{equation}
The incumbent records observed quality under the performed checks, without establishing correctness on untested conditions or identifying the most promising direction. An implementation error may obscure a useful idea, while an early gain may depend on narrow conditions. The leader agent weighs explanations, unresolved tests, and remaining resources before recommending continuation, diagnosis, redirection, or integration.


\section{Conditional Memory Sharing between Researcher Agents}
\label{sec:method}

Shared findings are useful only when recipients can assess their applicability. The leader-routed path in Figure~\ref{fig:areal-rsi-overview} preserves experimental context, limits the spread of unresolved problems, and supports local testing and corrective feedback.

\subsection{Experience with Sources and Applicability Conditions}
\label{sec:experience}

Shared memory comprises retrievable research records and communicated findings, distinct from local investigation histories and knowledge supplied by researcher harnesses. We represent the semantic content of an experience $e$ as:
\begin{equation}
 e=(d_e,f_e,c_e,u_e,\mathcal{F}_e),
 \label{eq:experience}
\end{equation}
where $d_e$ is the observed result, $f_e$ the proposed interpretation, $c_e$ the tested conditions, $u_e$ the remaining uncertainty, and $\mathcal{F}_e$ the supporting record references. Conditions describe inputs, acceptance criteria, and evaluation settings; references link to reports, artifacts, or execution traces. Separating observations from interpretations helps recipients assess whether the evidence supports the proposed explanation.

Each experience has a stable identifier, original source $o(e)$, current evidence source, and label $\kappa_t(e)\in\{\mathrm{problem},\mathrm{positive}\}$. A problem denotes an unresolved failure, risk, or difficulty; a positive experience denotes a finding the sender considers supported and usable. Identifiers, labels, sources, and references are structured metadata; the semantic fields in Equation~\ref{eq:experience} remain message content.
For example, a modification may reduce execution cost but fail a required check for some inputs. An omitted processing step is a possible explanation. Sharing both outcomes, the tested conditions, and that uncertainty supports diagnosis without prematurely rejecting or adopting the modification.

\subsection{Selective Propagation of Problems and Solutions}
\label{sec:propagation}

Broadcasting an unresolved failure can draw the entire team into the same uncertain diagnosis. A supported solution can instead remove a shared obstacle. \sys therefore restricts proactive problem routing while allowing positive findings to reach relevant researcher agents.

\textbf{Problem Routing.}
For experience $e$, let $\mathcal{R}^{\mathrm{problem}}_{e,t}$ contain researcher agents targeted by accepted, leader-initiated messages bearing the same identifier and the $\mathrm{problem}$ label through event $t$. The routing rule enforces
\begin{equation}
 \left|\mathcal{R}^{\mathrm{problem}}_{e,t}\setminus\{o(e)\}\right|\leq 1.
 \label{eq:problem-routing}
\end{equation}
The leader agent can recruit one additional researcher agent with relevant expertise. Further problem updates remain confined to that recipient and the original source under the same identifier.

\textbf{Positive Transfers.}
New evidence can support relabeling the experience as positive, allowing positive transfers without the additional-recipient bound. Let $\mathcal{F}^{\mathrm{new}}_e$ be the proposed references and $\mathcal{F}^{\mathrm{prev}}_e$ the references in the most recent registered update. Relabeling requires
\begin{equation}
 \mathcal{F}^{\mathrm{new}}_e\setminus
 \mathcal{F}^{\mathrm{prev}}_e\neq\varnothing,
 \label{eq:new-evidence}
\end{equation}
References must resolve within the campaign and include source-owned evidence for a researcher-agent source. These checks establish provenance and an additional reference, rather than certify the claimed solution. Interpretation, consistent classification, and semantic identity across retellings remain the leader agent's responsibility.

The bound governs proactive routing, not independent retrieval of public records or artifacts. Routine status summaries omit peer reports to avoid automatically broadcasting unresolved problems. Positive findings are also routed selectively according to relevance.

\subsection{Recipient Adaptation and Corrective Feedback}
\label{sec:adaptation}

Recipients compare shared findings with local conditions before adopting, adapting, rejecting, or deferring the advice. Candidates may be imported into a new subdirectory for local testing. Advice does not overwrite existing work or interrupt experiments.
Let $z_{n,i,t}$ be researcher agent $i$'s investigation state, $y_{n,i,t}$ the new local observations, $\mathcal{D}_{n,i,t}$ the delivered messages, and $\mathcal{P}_{n,i,t}$ the explicitly retrieved public records. Local reasoning updates the state as
\begin{equation}
 z_{n,i,t+1}=\operatorname{Update}_i
 \bigl(z_{n,i,t},y_{n,i,t},\mathcal{D}_{n,i,t},\mathcal{P}_{n,i,t}\bigr).
 \label{eq:recipient-update}
\end{equation}
The update incorporates available evidence without merging all team histories or requiring agreement. Applicability assessment and local verification are prescribed behavior, not an enforced adoption gate.

\textbf{Corrective Feedback.}
In the example above, a recipient might restore the omitted step and test whether the cost reduction survives while satisfying the required checks. Reporting the result and tested conditions under the same experience identifier allows the leader agent to qualify the finding and revise subsequent sharing. Accumulated findings also inform the harness revisions in \S\ref{sec:execution}.

\section{Experience-Driven Evolution of Researcher Harnesses}
\label{sec:execution}

Harness evolution turns completed investigations into procedures available to future teams. The across-campaign path in Figure~\ref{fig:areal-rsi-overview} links three steps: represent reusable research procedures, revise those procedures from evidence, and publish revisions for subsequent campaigns.

\subsection{Evolvable Research Orientation, Knowledge, and Skills}
\label{sec:harness}
\label{sec:inheritance}

A \emph{researcher harness} is the external configuration guiding an investigation:
\begin{equation}
 h=(p,\mathcal{K},\mathcal{S}),
 \label{eq:harness}
\end{equation}
Research orientation $p$ guides which hypotheses to pursue. Knowledge $\mathcal{K}$ records findings, applicability conditions, and failure boundaries. Skills $\mathcal{S}$ specify repeatable procedures, including inputs, steps, validation criteria, and supporting scripts. Together, these components influence what a researcher agent investigates, what evidence is available, and how experiments are conducted.

For the example in \S\ref{sec:experience}, knowledge can retain the conditions under which a modification failed, while a skill adds targeted checks before full evaluation. The reusable outcome is a better-informed procedure for investigating future candidates.

The implementation packages the harness as a \emph{profile}, supplied through initial context and workspace resources. Revisions affect these assets while model weights, tool permissions, orchestration, and the leader agent's policy remain unchanged.

\subsection{From Experimental Experience to Procedural Revision}
\label{sec:review}
\label{sec:memoryevolution}
\label{sec:skillevolution}

After exploration and applicable final verification, the leader agent enters a separate review session. The review examines the task, selected harnesses, reports, and retrievable campaign records, with instructions to propose reusable revisions without new experiments.

\textbf{Evidence-Based Revision.}
Let $\mathcal{H}_n=\{h_{n,i}:1\leq i\leq K\}$ denote the distinct harnesses used in campaign $n$, $\mathcal{E}_n$ the completed public evidence, and $v_n$ the finalization feedback. The leader agent's review policy $\pi_{\mathrm{review}}$ proposes edits $\Delta_n$:
\begin{equation}
 \Delta_n\sim\pi_{\mathrm{review}}
 (\,\cdot\mid q_n,\mathcal{H}_n,\mathcal{E}_n,v_n),
 \qquad
 h'_j=\operatorname{Apply}(h_j,\Delta_{n,j}).
 \label{eq:harness-review}
\end{equation}
Here $j$ identifies a selected harness and $\Delta_{n,j}$ contains the corresponding changes. Evidence is accessed through retrieval; an empty proposal is valid when no reusable revision is justified.

Review asks which orientations helped, which findings generalize beyond the task, and which repeated operations can become skills. Proposals should retain applicability conditions, uncertainty, and contradictory evidence. Instructions exclude complete task answers and hidden-test information from reusable assets; this is a semantic requirement rather than a formal guarantee.

Each proposal supplies a rationale, same-campaign evidence references, and asset additions or replacements. Updates preserve unspecified assets and remain restricted to selected harness identities. Publication checks references, structure, and required resources; acceptance requires neither script execution nor demonstrated improvement on future tasks.

\subsection{Reuse and Specialization in Subsequent Campaigns}
\label{sec:reuse}
\label{sec:lineage}

\textbf{Versioned Publication.}
Accepted revisions update the shared library used to initialize new campaigns. Let $\mathcal{L}^{\mathrm{cur}}$ be the library before publication and $\mathcal{H}'_n$ the proposed revised harnesses. Replacement requires admissible revisions whose current versions still match those reviewed:
\begin{equation}
 \mathcal{L}^{+}=\begin{cases}
 \operatorname{Replace}(\mathcal{L}^{\mathrm{cur}},\mathcal{H}'_n),
 & \text{if admissible and version-consistent},\\
 \mathcal{L}^{\mathrm{cur}}, & \text{otherwise}.
 \end{cases}
 \label{eq:publication}
\end{equation}
The version check prevents stale reviews from overwriting newer changes. Failed or conflicting proposals leave the library unchanged.

For campaigns created after the preceding review, the adaptation loop is
\begin{equation}
 \overline{\mathcal{L}}_n
 \xrightarrow{\ \mathrm{select}\ }\mathcal{H}_n
 \xrightarrow{\ \mathrm{research\ and\ share}\ }\mathcal{E}_n
 \xrightarrow{\ \mathrm{review\ and\ publish}\ }\overline{\mathcal{L}}_{n+1}.
 \label{eq:adaptation-loop}
\end{equation}
Earlier campaigns retain their snapshots. New teams select revised harnesses through Equation~\ref{eq:composition}, without inheriting unrestricted access to past traces or complete answers. Specialization combines revisions to existing harness identities with task-conditioned selection; the update operator does not create new identities.

\textbf{Transfer Assessment.}
A harness revision should improve future investigations, whereas an artifact is evaluated on the current task. Assessing transfer therefore requires later campaigns with declared task splits, library versions, and feedback permissions. Source-campaign finalization feedback may inform review, but target held-out outcomes must remain outside updates used to solve that target.

\textbf{Resource Accounting.}
Review also consumes resources. With $\cost^{\mathrm{campaign}}_n$ covering coordination, researcher-agent execution, and verification, and $\cost^{\mathrm{review}}_n$ covering post-campaign review, total expenditure is
\begin{equation}
 \cost^{\mathrm{total}}_n=
 \cost^{\mathrm{campaign}}_n+\cost^{\mathrm{review}}_n.
 \label{eq:total-cost}
\end{equation}
These vectors contain additive costs such as tokens, model charges, and experiment compute; elapsed wall time is reported separately. The review phase has a separate allowance and must be included when measuring the benefit of reuse.

\Needspace{8\baselineskip}
\section{Evaluation}
\label{sec:evaluation}

We evaluate whether \sys translates parallel researcher-agent computation into better research outcomes. The evaluation addresses two research questions: (\underline{\textbf{RQ1}}) \textit{How does \sys's end-to-end performance compare with the popular \textsc{Codex-Goal} baseline?} (Section~\ref{sec:eval-rq1}); and (\underline{\textbf{RQ2}}) \textit{How do the proposed mechanisms improve the scalability of parallel research?} (Section~\ref{sec:eval-rq2}).
We first specify the tasks and evaluation settings (Section~\ref{sec:eval-tasks}), then report the results for the research questions.

\subsection{Experimental Setup}
\label{sec:eval-tasks}
\textbf{Tasks.}
We evaluate three task families: downstream Terminus harness optimization on Terminal Bench 2.1 (89 tasks, with the task-solving model fixed); GPU kernel optimization on seven KernelBench tasks~\citep{kernelbench}; and four RSI Bench research tasks covering on-policy self-distillation, jailbreak robustness, agent-swarm optimization, and nanoGPT data curation. The KernelBench tasks are L1P63, L1P77, L2P14, L2P71, L3P10, L3P19, and L3P28.

\textbf{Methods and Resources.}
We compare \sys and \textsc{Codex-Goal} using GPT-5.6-Luna and GPT-6-Astra as research backbones. KernelBench allows one effective hour for \textsc{Codex-Goal} and for each \sys researcher agent; RSI Bench allows two effective hours for \textsc{Codex-Goal} and for each \sys researcher agent. RQ2 compares independent four-agent exploration, experience sharing, and sharing with specialized researcher roles.

\textbf{Evaluation Protocol.}
For Terminal Bench, \sys iterates and selects on smaller training and validation subsets, whereas \textsc{Codex-Goal} evaluates the full task set each round. Both selected harnesses are independently reevaluated on all 89 tasks. KernelBench specifies seed 137 for independent final reevaluation. Each candidate was evaluated on a single,
exclusively allocated GPU, with warm-up
followed by 100 repeated timing measurements
using CUDA events.

\textbf{Metrics.}
Terminal Bench reports resolution rate. KernelBench reports arithmetic mean speedup across seven tasks, assigning zero to unsuccessful tasks; speedup is reference runtime divided by candidate runtime. RSI Bench uses the task-specific metrics in Table~\ref{tab:main-results}, with larger values preferred except for perplexity. Configuration comparisons report kernel latency in milliseconds. \emph{Fail} means no correct final candidate. Comparisons are descriptive; repeated-run uncertainty is unavailable.

\Needspace{6\baselineskip}
\subsection{End-to-End Performance Compared with \textsc{Codex-Goal}}
\label{sec:eval-rq1}
\textbf{Comparison.}
Table~\ref{tab:main-results} addresses RQ1 within each research backbone. \sys improves all six reported metrics with both Luna and Astra, giving 12 favorable comparisons. The consistent direction spans downstream harnesses, kernels, and AI research, while the size of the gains varies by task and backbone.

\begin{table}[!htbp]
\centering
\caption{End-to-end research outcomes under the reported resource allocations. Bold identifies the better available numerical value within each backbone. Arrows indicate the preferred direction.}
\vspace{0.5em}
\label{tab:main-results}
\begingroup
\small
\setlength{\tabcolsep}{3pt}
\renewcommand{\arraystretch}{1.2}
\begin{tabular*}{\textwidth}{@{\extracolsep{\fill}}lrrrrrr@{}}
\toprule
& \multicolumn{1}{c}{Terminal Bench 2.1}
& \multicolumn{1}{c}{KernelBench}
& \multicolumn{4}{c}{RSI Bench} \\
\cmidrule(lr){2-2}\cmidrule(lr){3-3}\cmidrule(l){4-7}
Method
& \multicolumn{1}{c}{\shortstack{Resolved\\(\%) $\uparrow$}}
& \multicolumn{1}{c}{\shortstack{Mean speedup\\($\times$) $\uparrow$}}
& \multicolumn{1}{c}{\shortstack{Self-distill.\\score (\%) $\uparrow$}}
& \multicolumn{1}{c}{\shortstack{Jailbreak\\score $\uparrow$}}
& \multicolumn{1}{c}{\shortstack{Swarm\\accuracy (\%) $\uparrow$}}
& \multicolumn{1}{c}{\shortstack{Data curation\\PPL $\downarrow$}} \\
\midrule
\multicolumn{7}{@{}l}{\textbf{GPT-5.6-Luna}} \\
\textsc{Codex-Goal} & 75.3 & 1.74 & 44.17 & 11.251 & 53.33 & 571.534 \\
\sys & \textbf{77.5} & \textbf{2.30} & \textbf{55.83} & \textbf{35.544} & \textbf{66.67} & \textbf{422.837} \\
\addlinespace[4pt]
\multicolumn{7}{@{}l}{\textbf{GPT-6-Astra}} \\
\textsc{Codex-Goal} & 85.4 & 6.28 & 51.94 & 81.142 & 60.00 & 377.351 \\
\sys & \textbf{87.6} & \textbf{8.59} & \textbf{57.50} & \textbf{81.261} & \textbf{73.33} & \textbf{340.523} \\

\bottomrule
\end{tabular*}
\endgroup
\end{table}

\textbf{Downstream Harness Optimization.}
Terminal Bench resolution increases from 75.3\% to 77.5\% with Luna and from 85.4\% to 87.6\% with Astra, a gain of 2.2 percentage points for each backbone. Because the task-solving model is fixed, these gains reflect changes to the downstream harness produced by each research system. This experiment is distinct from evolving the researcher harnesses in Section~\ref{sec:execution}.

\textbf{GPU Kernel Optimization.}
Mean speedup across seven KernelBench tasks increases from $1.74\times$ to $2.30\times$ with Luna and from $6.28\times$ to $8.59\times$ with Astra, relative gains of 32.2\% and 36.8\%. Improvements with both backbones show that the advantage extends beyond one research model. Because unsuccessful tasks contribute zero, the aggregate combines candidate correctness and execution performance; the summary does not separate their contributions.

\textbf{AI Research Tasks.}
Self-distillation scores improve by 11.66 percentage points with Luna and 5.56 with Astra. Swarm accuracy improves by 13.34 and 13.33 percentage points, respectively, while data-curation perplexity decreases by 26.0\% and 9.8\%. Jailbreak scores improve by 24.293 with Luna but only 0.119 with Astra. These gains span all four research objectives, although the Astra jailbreak difference is small.

\vspace{0.5em}
\begin{rqsummary}{RQ1 Key Takeaway}
Under the reported resource allocations, \sys outperforms \textsc{Codex-Goal} on all six metrics with both backbones. The results support end-to-end effectiveness across tasks, without establishing equal-compute efficiency.
\end{rqsummary}

\subsection{Comparison of Parallel Research Configurations}
\label{sec:eval-rq2}
\textbf{Comparison.}
Table~\ref{tab:parallel-configurations} addresses RQ2 by comparing independent four-agent exploration, experience sharing, and sharing with specialized researcher roles. The comparison covers four KernelBench tasks and two backbones. All three configurations produce valid candidates for the same seven task--backbone pairs; Luna's L3P10 fails under every configuration.

\begin{table}[!htbp]
\centering
\caption{Comparison of parallel research configurations on KernelBench. Entries are final candidate latencies in milliseconds ($\downarrow$). Bold marks the lowest latency for each task--backbone pair. Fail means no correct final candidate.}
\vspace{0.5em}
\label{tab:parallel-configurations}
\begingroup
\small
\setlength{\tabcolsep}{3.5pt}
\renewcommand{\arraystretch}{1.18}
\begin{tabularx}{\textwidth}{@{}l*{8}{>{\raggedleft\arraybackslash}X}@{}}
\toprule
& \multicolumn{4}{c}{\textbf{GPT-6-Astra}}
& \multicolumn{4}{c}{\textbf{GPT-5.6-Luna}} \\
\cmidrule(lr){2-5}\cmidrule(l){6-9}
Configuration & L1P63 & L2P71 & L3P10 & L3P19 & L1P63 & L2P71 & L3P10 & L3P19 \\
\midrule
Independent (four agents) & 5.200 & 0.338 & 8.815 & 0.482 & 11.944 & 0.523 & \textit{Fail} & 1.766 \\
Experience sharing & 4.731 & 0.294 & 7.379 & 0.479 & 15.633 & 0.593 & \textit{Fail} & 1.579 \\
Sharing + researcher roles & \textbf{4.665} & \textbf{0.251} & \textbf{6.395} & \textbf{0.473} & \textbf{10.498} & \textbf{0.392} & \textit{Fail} & \textbf{1.407} \\

\bottomrule
\end{tabularx}
\endgroup
\end{table}

\textbf{Sharing Alone.}
Experience sharing lowers latency in five of seven comparable pairs: all four Astra tasks and one of the three successful Luna tasks. However, Luna's L1P63 latency rises from 11.944 to 15.633\,ms (30.9\%), and L2P71 rises from 0.523 to 0.593\,ms (13.4\%). The sharing configuration therefore yields better or worse final artifacts depending on the task and backbone.

\textbf{Sharing with Specialization.}
The combined configuration achieves the lowest latency in all seven comparable pairs, reducing latency by 1.9--27.5\% relative to independent exploration and by 1.3--33.9\% relative to sharing alone. On Astra's L3P10, the three configurations yield 8.815, 7.379, and 6.395\,ms, respectively. On Luna's L2P71, the combined configuration records 0.392\,ms despite the sharing-only regression, 25.0\% below independent exploration and 33.9\% below sharing alone. The combined configuration records lower latency both where sharing alone improves on independent exploration and where sharing alone performs worse.

\textbf{Discussion.}
The pattern favors organizing information reuse together with differentiated investigative roles, rather than relying on information exchange alone. This interpretation is consistent with task-conditioned harness selection and conditional memory sharing, but the comparisons do not isolate individual mechanism effects or measure research-direction diversity. The study evaluates configurations at the reported parallelism; scaling curves, matched-compute efficiency, and gains from cross-campaign harness evolution remain unmeasured.

\begin{rqsummary}{RQ2 Key Takeaway}
Sharing with specialized researcher roles gives the best result in all seven comparable task-backbone pairs, whereas sharing alone has mixed outcomes. The evidence favors the combined configuration at the evaluated parallelism.
\end{rqsummary}
\Needspace{9\baselineskip}
\section{Related Work}
\label{sec:related}

\textbf{RSI for Agent Evolution.} Agent evolution ranges from optimizing agent designs and prompts to revising the improvement process. ADAS searches over agent designs~\citep{adas}, the Darwin G\"odel Machine evolves coding-agent programs~\citep{dgm}, and GEPA evolves prompts through reflection~\citep{gepa}. DGM keeps pretrained model weights fixed, locating improvement in the agent's executable procedures~\citep{dgm}. Hyperagents makes the self-modification procedure editable~\citep{hyperagents}, while Dream-RSI refines exploration policies using replay over historical discovery trees~\citep{dreamrsi}. $\tau^\tau$-Bench evaluates agent construction and subsequent operating cost~\citep{hypertau}.

\textbf{Autonomous Research and Agentic Search.} AIDE and AI Scientist-v2 organize program development and experimentation through tree search~\citep{aide,aiscientistv2}; AlphaEvolve and Loreley maintain evolutionary archives of programs and repository states, respectively~\citep{alphaevolve,loreley}. Retaining earlier candidates supports both continued refinement and exploration from alternative starting points. Population Based Training reallocates workers through parameter inheritance and hyperparameter perturbation~\citep{pbt}. Fluid search allocates effort and reseeds stalled research chains~\citep{efficiencymatters}, ExTS balances expansion and refinement~\citep{exts}, and Research Preference Models screen proposals before expensive evaluation~\citep{researchpreferences}.
RE-Bench and MLE-bench evaluate research engineering~\citep{rebench,mlebench}; KernelBench, Scale RSI Bench, and AutoWorldModel-Bench provide tasks in kernel optimization, AI research, and world modeling~\citep{kernelbench,scalersi,autoworldmodel}.

\textbf{Agent Coordination and Memory Sharing.}
AI co-scientist and Arbor coordinate specialized agents and hypothesis-driven investigations~\citep{coscientist,arbor}; EurekAgent supports collaboration through filesystem and Git artifacts~\citep{eurekagent}, and BusMA combines worker-local memory with message-driven coordination~\citep{busma}.
Persistent records allow accumulated evidence to remain accessible after an individual agent session ends. Stateful ReAct maintains typed persistent state~\citep{statefulreact}, G-Memory retrieves hierarchical experience~\citep{gmemory}, MemCon controls memory operations~\citep{memcon}, and CoMem combines private and curated collective memory~\citep{comem}. Materials-science memory and Repo-To-Skill organize reusable knowledge and executable skills~\citep{materials_memory,repotoskill}. EdgeBench and Beyond Final Scores study the effects of retained experience on continued research~\citep{edgebench,beyondfinal}.
Independent-agent aggregation provides a scaling baseline~\citep{agentforest}, while coordination benefits depend on task structure and overhead~\citep{scalingagents}. Sharing can also homogenize proposals~\citep{interactiontax} or propagate evaluator exploits~\citep{swarmcheating}.

\section{Conclusion}
\label{sec:conclusion}

In this paper, we introduced \sys, which formulates test-time RSI scaling as the organization of persistent investigations and reusable research procedures. The framework combines conditional memory sharing within campaigns with evidence-driven harness evolution across campaigns. Under the evaluated resource allocations, \sys improves Terminal Bench resolution by 2.2 percentage points and mean KernelBench speedup by 32.2--36.8\% over \textsc{Codex-Goal}. Combining experience sharing with specialized researcher roles achieves 1.9--27.5\% lower kernel latency than independent exploration and 1.3--33.9\% lower latency than sharing alone across all seven comparable task--backbone pairs. These results support coordinated parallel research, while matched-compute scaling and the benefit of cross-campaign harness evolution remain to be established.
\FloatBarrier

\bibliographystyle{unsrtnat}
\bibliography{references}
\clearpage
\end{document}